%% file: aaai23.tex
\documentclass[letterpaper]{article} 
\usepackage{aaai23}  
\usepackage{times}  
\usepackage{helvet}  
\usepackage{courier}  
\usepackage[hyphens]{url}  
\usepackage{graphicx} 
\usepackage{natbib}  
\usepackage{caption} 
\usepackage{algorithm}
\usepackage{algorithmic}
\usepackage{epsfig}
\usepackage{amsmath}
\usepackage{amssymb}
\usepackage{amsfonts}       
\usepackage{tikz}
\usepackage{comment}
\usepackage{color}
\usepackage{multirow}
\usepackage{booktabs}       
\usepackage{dsfont}
\usepackage{bm}  
\usepackage{xspace}
\usepackage{colortbl}
\usepackage{enumitem}
\usepackage{placeins}
\usepackage{footnote}
\usepackage[title]{appendix}
\usepackage{kotex}
\usepackage{bbding}
\usepackage{microtype}      
\usepackage{nicefrac}       
\usepackage{subcaption, overpic, textpos}
\usepackage{wrapfig}
\usepackage{tabulary}
\usepackage{relsize}

\include{math_commands}
\definecolor{Gray}{gray}{0.9}

\newcommand{\eqnspace}{
\setlength{\abovedisplayskip}{5pt plus 1.0pt minus 1.0pt}
\setlength{\belowdisplayskip}{5pt plus 1.0pt minus 1.0pt}
\setlength{\abovedisplayshortskip}{0.0pt plus 3.0pt}
\setlength{\belowdisplayshortskip}{6.0pt plus 3.0pt minus 3.0pt}}

\newlength\savewidth\newcommand\shline{\noalign{\global\savewidth\arrayrulewidth
  \global\arrayrulewidth 1pt}\hline\noalign{\global\arrayrulewidth\savewidth}}
\newcommand{\tablestyle}[2]{\setlength{\tabcolsep}{#1}\renewcommand{\arraystretch}{#2}\centering\footnotesize}

\newcolumntype{x}[1]{>{\centering\arraybackslash}p{#1pt}}
\newcolumntype{y}[1]{>{\raggedright\arraybackslash}p{#1pt}}
\newcolumntype{z}[1]{>{\raggedleft\arraybackslash}p{#1pt}}

\newcommand{\eqnsm}[2]{\begin{equation}\label{eq:#1}#2\end{equation}}

\newlength\secmargin
\newlength\subsecmargin
\newlength\paramargin
\newlength\abovetabcapmargin
\newlength\belowtabcapmargin
\newlength\abovefigcapmargin
\newlength\belowfigcapmargin

\usepackage{pifont}
\newcommand{\cmark}{\ding{51}}
\newcommand{\xmark}{\ding{55}}

\makeatletter
\DeclareRobustCommand\onedot{\futurelet\@let@token\@onedot}
\def\@onedot{\ifx\@let@token.\else.\null\fi\xspace}

\def\eg{\emph{e.g}\onedot} 
\def\ie{\emph{i.e}\onedot} 
 
\def\etc{\emph{etc}\onedot} \def\vs{\emph{vs}\onedot}
\def\wrt{w.r.t\onedot} 
 
\def\aka{a.k.a\onedot} \def\etal{\emph{et al}\onedot}
\makeatother

\newcommand*\rot{\rotatebox{90}}

\newcommand{\ours}{\mbox{Action\textsc{mae}}\xspace}

\usepackage{newfloat}
\usepackage{listings}
\DeclareCaptionStyle{ruled}{labelfont=normalfont,labelsep=colon,strut=off}
\floatstyle{ruled}
\newfloat{listing}{tb}{lst}{}
\floatname{listing}{Listing}

\definecolor{Green}{rgb}{0.2, 0.7, 0.1}
\definecolor{Orange}{rgb}{0.8, 0.5, 0.2}
\definecolor{citecolor}{HTML}{0071BC}
\definecolor{linkcolor}{HTML}{ED1C24}
\definecolor{plus}{HTML}{0071bc}
\definecolor{minus}{RGB}{153,10,10}
\definecolor{rgb_red}{HTML}{BB3E03}
\definecolor{depth_teal}{HTML}{0A9396}
\definecolor{ir_yellow}{HTML}{EE9B00}

\newcommand\crgb[1]{\textcolor{rgb_red}{#1}}
\newcommand\cdepth[1]{\textcolor{depth_teal}{#1}}
\newcommand\cir[1]{\textcolor{ir_yellow}{#1}}
\newcommand\rgb{\crgb{RGB}\xspace}
\newcommand\depth{\cdepth{Depth}\xspace}
\newcommand\ir{\cir{IR}\xspace}
\newcommand\R{\crgb{R}\xspace}
\newcommand\D{\cdepth{D}\xspace}
\newcommand\I{\cir{I}\xspace}
\newcommand{\pacc}[1]{{\bf \fontsize{8}{42}\selectfont \color{plus!80} #1}}
\newcommand{\macc}[1]{{\bf \fontsize{8}{42}\selectfont \color{minus!70} #1}}
\newcommand{\up}{\bf \fontsize{10}{42} \color{plus}{$\uparrow$}}
\newcommand{\down}{\bf \fontsize{10}{42}\selectfont \color{minus}{$\downarrow$}}
\usepackage[pagebackref=true,breaklinks=true,colorlinks,citecolor=citecolor,linkcolor=linkcolor,bookmarks=false]{hyperref}
\usepackage[capitalize]{cleveref}
\crefname{section}{Sec.}{Secs.}
\Crefname{section}{Section}{Sections}
\Crefname{table}{Table}{Tables}
\crefname{table}{Tab.}{Tabs.}

\title{Towards Good Practices for Missing Modality Robust Action Recognition}
\author{
    Sangmin Woo,
    Sumin Lee,
    Yeonju Park,
    Muhammad Adi Nugroho,
    Changick Kim
}
\affiliations{
    Korea Advanced Institue of Science and Technology (KAIST)\\
    \{smwoo95, suminlee94, yeonju29, madin, changick\}@kaist.ac.kr
}

\begin{document}
\maketitle
\begin{abstract}
Standard multi-modal models assume the use of the same modalities in training and inference stages.
However, in practice, the environment in which multi-modal models operate may not satisfy such assumption.
As such, their performances degrade drastically if any modality is missing in the inference stage.
We ask: \textit{how can we train a model that is robust to missing modalities?}
This paper seeks a set of good practices for multi-modal action recognition, with a particular interest in circumstances where some modalities are not available at an inference time.
First, we study how to effectively regularize the model during training (\eg, data augmentation).
Second, we investigate on fusion methods for robustness to missing modalities: we find that transformer-based fusion shows better robustness for missing modality than summation or concatenation.
Third, we propose a simple modular network, \ours, which learns missing modality predictive coding by randomly dropping modality features and tries to reconstruct them with the remaining modality features.
Coupling these good practices, we build a model that is not only effective in multi-modal action recognition but also robust to modality missing.
Our model achieves the state-of-the-arts on multiple benchmarks and maintains competitive performances even in missing modality scenarios.
Codes are available at \url{https://github.com/sangminwoo/ActionMAE}.
\end{abstract}

\eqnspace
\FloatBarrier
\input{01introduction}
\input{02preliminary}
\input{03method}
\input{04experiments}
\input{05related_work}
\input{06discussion}
\input{07acknowledgement}

\input{08appendix} 
\FloatBarrier

\bibliography{aaai23}

\end{document}

%% file: 01introduction.tex
\vspace{\secmargin}\section{Introduction}
\begin{figure}[t!]
    \centering
    \resizebox{\linewidth}{!}{
        \includegraphics[width=\linewidth]{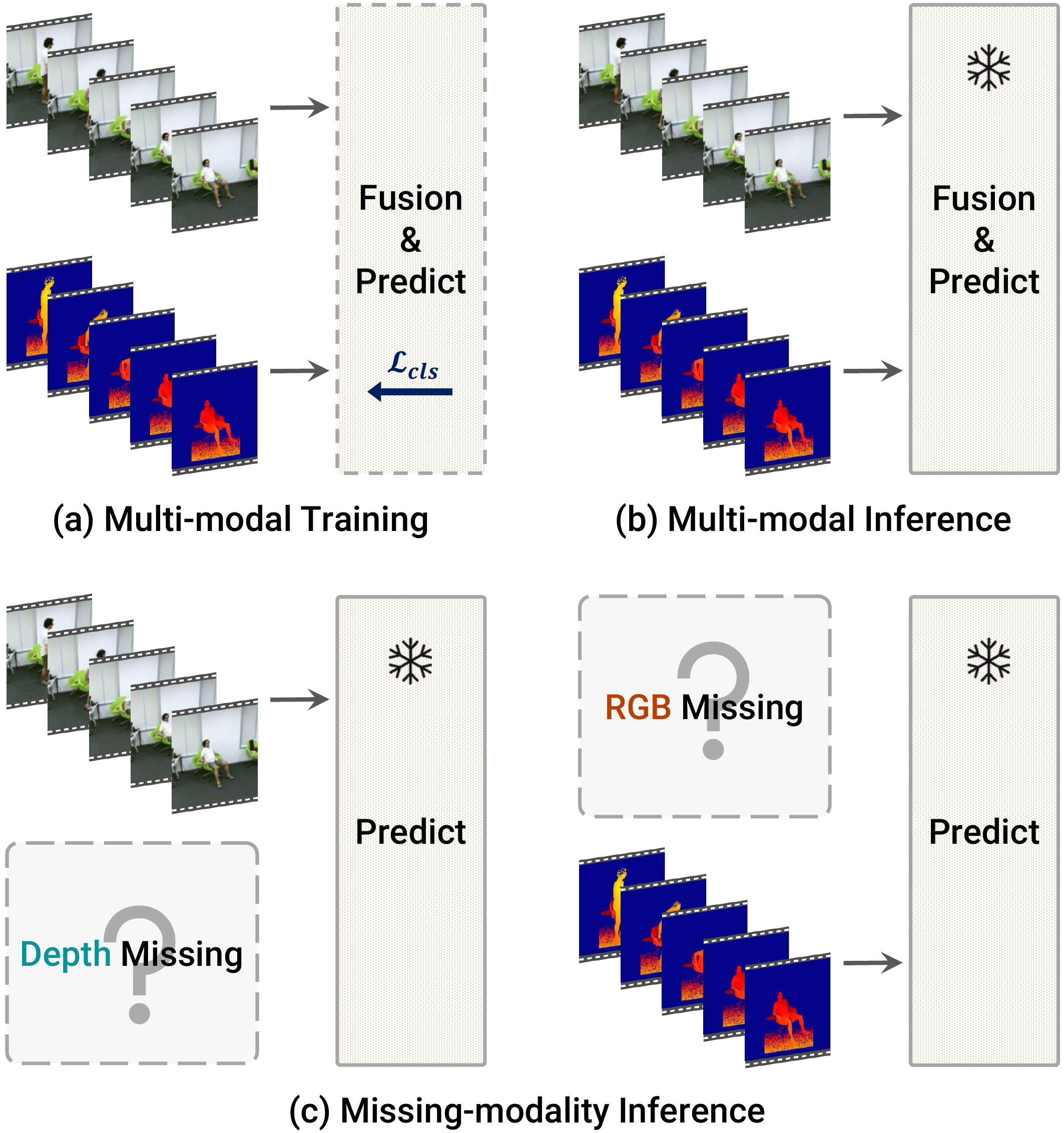}
    }
    \vspace{\abovefigcapmargin}
    \vspace{-3mm}
    \caption{
        {Action recognition with missing modality.}
        Standard multi-modal action recognition assumes that the modalities used in the training stage are complete at inference time: (a) $\rightarrow$ (b).
        This paper addresses the action recognition problem in situations where such assumption is not established, \ie, when modalities are incomplete at inference time: (a) $\rightarrow$ (c).
        Our goal is to maintain performance in the absence of any input modality.
        \SnowflakeChevron~indicates the weight-frozen.
    }
    \label{fig:missing_modality}
    \vspace{\belowfigcapmargin}
    \vspace{-1mm}
\end{figure}

This study aims to answer the underlying question about multi-modal learning for action recognition in practical situations: \textit{How can we train a model that is robust to missing modalities?}
Typical multi-modal models assume complete modalities in both training and inference phases~\cite{bruce2021multimodal,bruce2022mmnet}.
In reality, however, the multi-modal system may be unable to access particular modalities during the inference phase, despite being able to access all modalities reliably during the training phase (see~\Cref{fig:missing_modality}).
There are a number of potential causes for such circumstances, including malfunctioning sensors, high data acquisition costs, inaccessibility due to security or privacy concerns, self-deficiencies, \etc.
This situation affects the reliability, accuracy, and safety of the model in real-world applications.
Considering the autonomous driving situation, an error caused by insufficient use of the sensors due to factors such as inclement weather would be life-threatening.
As such, the missing modality problem is critical when multi-modal models are employed in practice.

\input{tabs/missing_performance}

\begin{figure}[t!]
    \centering
    \resizebox{0.92\linewidth}{!}{
        \includegraphics[width=\linewidth]{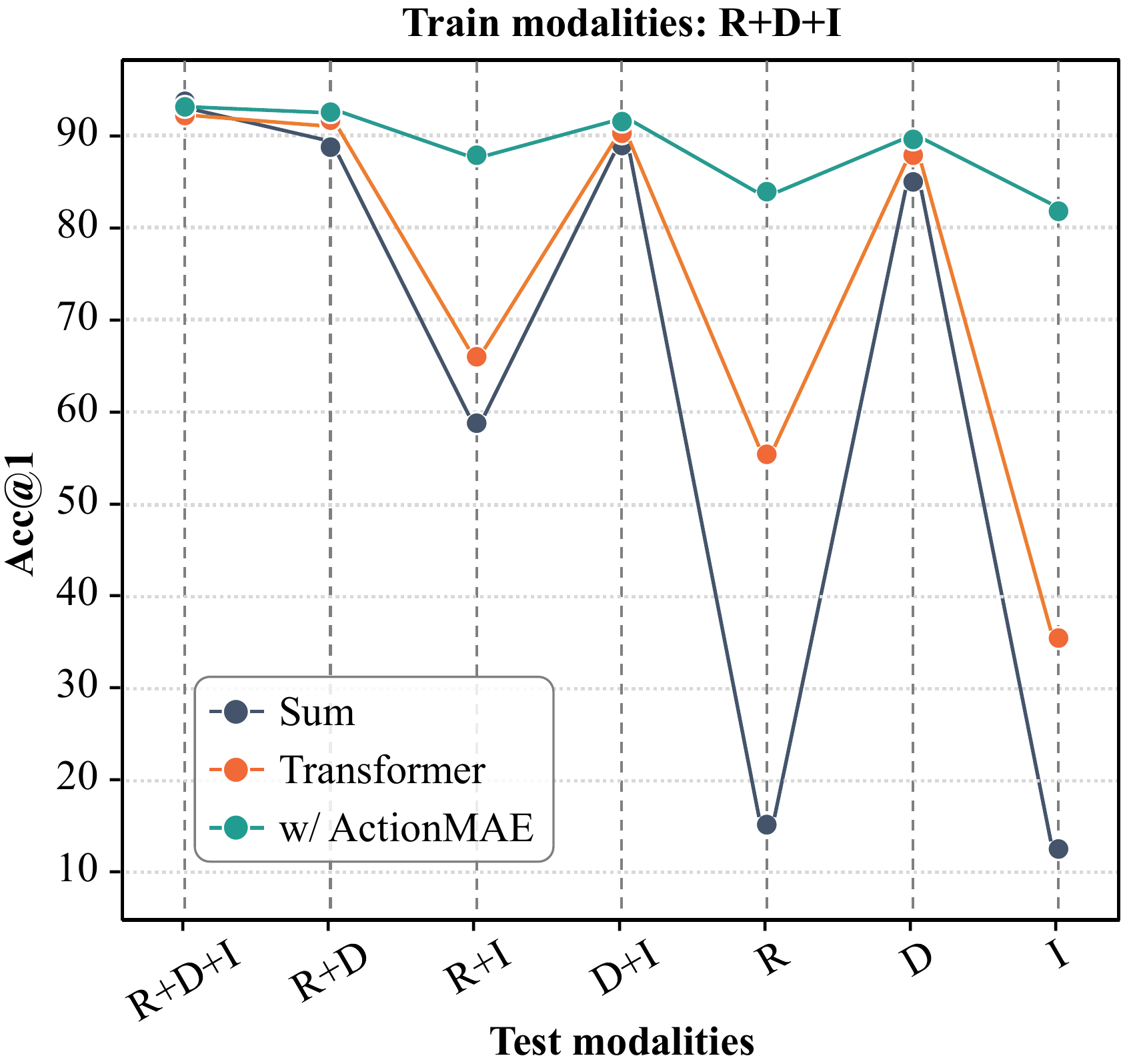}
    }
    \vspace{\abovefigcapmargin}
    \caption{
        {Towards missing modality robust action recognition.}
        \textbf{(i)} Transformer fusion is robust than sum fusion.
        \textbf{(ii)} In contrast to conventional approaches, our \ours well preserves the multi-modal performance even with incomplete modalities.
        All models are trained with {\R}+{\D}+{\I} modalities and tested on NTU RGB+D 60~\cite{shahroudy2016ntu}.
    }
    \label{fig:teaser}
    \vspace{\belowfigcapmargin}
\end{figure}

To address the missing modality scenarios, we require a robust model that operates reliably even when certain modalities are unavailable in multi-modal systems.
However, conventional multi-modal models are learnt under the premise that all modalities exist, so that they cannot achieve sufficient performances when directly applied to missing modality settings.
\Cref{tab:missing_performance} shows how much the conventional multi-modal models suffer from missing modality.
Our goal is to achieve on-par results as when all modalities are available, even if any of the input modality is missing.
We achieve this by designing a set of good practices for missing modality scenarios in multi-modal action recognition (see~\Cref{fig:teaser}):
\noindent\textbf{(i)} As a preliminary step, we first demonstrate a recipe of how to effectively build a multi-modal model with several training ingredients such as architecture, data augmentation, and regularization.
\noindent\textbf{(ii)} We observe that the fusion method matters for robustness to missing modality.
We compare the three most widely used methods in the literature among the numerous methods for multi-modal fusion: summation, concatenation, and transformer.
Among them, we find that transformer-based fusion method is the most robust against missing modality.
\noindent\textbf{(iii)} We propose Action Masked Auto Encoder (\ours), a simple modular network that learns missing modality predictive coding.
The model with \ours is optimized under the two objectives. It jointly learns to reconstruct the missing modality features and to predict the correct action label during training.
We can plug-and-play \ours on top of standard space-time encoders.
Despite its simplicity, \ours achieves a win-win situation: it not only regularizes the multi-modal model to achieve better accuracy in complete modality cases, but also maintains accuracy in missing modality cases. 

We report our results on four challenging action recognition benchmarks: NTU RGB+D 60~\cite{shahroudy2016ntu}, NTU RGB+D 120~\cite{liu2019ntu}, NW-UCLA~\cite{wang2014cross}, and UWA3D~\cite{rahmani2016histogram}.
We set new state-of-the-art results in both complete and missing modality settings.
Furthermore, we observe some intriguing properties of \ours from experiments, including the following:
\textbf{(i)} Missing modality reconstruction yields a nontrivial and meaningful self-supervision. As such, the results with reconstructed features are superior than those with the original features extracted directly from space-time encoder.
\textbf{(ii)} \ours effectively regularizes the multi-modal model, hence the model equipped with \ours achieves performance gain over the vanilla model when using complete modalities. 
\textbf{(iii)} Moreover, \ours alleviates the bias of the multi-modal model toward the dominant modality (\ie, the most contributing modality to the learning objective).
\textbf{(iv)} The robustness of \ours is agnostic to the type or the number of missing modalities. This suggests that it can preserve accuracy even in real-world environments where we do not know which modality might be missing.


%% file: tabs/missing_performance.tex
\begin{table}[t!]
\tablestyle{2pt}{1.1}
\begin{tabular}{y{65}|x{50}x{45}x{28}x{28}}
Dataset & Train modal$^{*}$ & Test modal & Acc.$^{\dagger}$ & $\triangle$ (\%p) \\
\shline
\multirow{3}{*}{NTU-RGB+D 60}   & \cellcolor{Gray} {\R}+{\D}+{\I} & \cellcolor{Gray} {\R}+{\D}+{\I} & \cellcolor{Gray} 93.3\%  & \cellcolor{Gray} - \\
                                & {\R}+{\D}+{\I} & {\R}+{\I} &   59.4\%    &~\macc{$-$33.9} \\
                                & {\R}+{\D}+{\I} & {\R}   &   15.6\%    &~\macc{$-$77.7} \\
\hline
\multirow{3}{*}{NW-UCLA}        & \cellcolor{Gray} {\R}+{\D}   & \cellcolor{Gray} {\R}+{\D} & \cellcolor{Gray} 91.9\%  & \cellcolor{Gray} - \\
                                & {\R}+{\D}   & {\R}   &   53.6\%    &~\macc{$-$38.3} \\
                                & {\R}+{\D}   & {\D}   &   70.3\%    &~\macc{$-$21.6} \\
\multicolumn{5}{c}{*\textbf{{\R}}: RGB, \textbf{{\D}}: depth, \textbf{{\I}}: infrared. $\dagger$~Results based on sum fusion.}
\end{tabular}
\vspace{\abovetabcapmargin}
\caption{
{Multi-modal models suffer severely in missing modality scenarios.}
The rows marked in \colorbox{Gray}{gray} use the same modalities in training and inference stages, which are standard settings of multi-modal inference.
We observe significant performance drop when the modality is removed one by one from the complete modality setup in inference stage.
}
\vspace{\belowtabcapmargin}
\label{tab:missing_performance}
\end{table}

%% file: 02preliminary.tex
\vspace{\secmargin}\section{Preliminary: Multi-modal Action Classifier}
We first study how to design a strong multi-modal action recognition model with several design choices.

\vspace{\subsecmargin}\subsection{Architecture}
The standard architecture of multi-modal action recognition is in a form of space-time encoder followed by a fusion unit.
We seek the optimal space and time encoders for multi-modal action recognition with the fusion unit set to a simple summation by default.
\Cref{tab:architecture} compares the accuracies of the simplest combinations among the numerous candidates: R(2+1)D~\cite{tran2018closer}, ViT~\cite{dosovitskiy2020image}, ResNet~\cite{he2016deep}, Transformer~\cite{vaswani2017attention}.
\input{tabs/architecture}

We choose ResNet34 and Transformer as our space and time encoders, respectively.
The procedure of space-time encoding is formalized as follows.
We sample a sequence of frames $[f_i^m]_{i=1}^{T}$ ($f_i^m \in \mathbb{R}^{C_0 \times H \times W}$) from a video $\mathcal{V}^m$, where $i$ indicates the frame index and $m$ indicates the modality.
In practice, we set $T=16$, $C_0=3$, $H=W=224$, and $m \in {\rm \{R, D, I\}}$.
A frame of modality $m$ is processed with a modality-specific space encoder $\mathcal{E}^{m}$ followed by an average pooling:
\eqnsm{space_encoder}{x_i^m = \texttt{avgpool}(\mathcal{E}^{m}(f_i^m))\,.}

The output feature is $x^m_i \in \mathbb{R}^{C}$, where $C=512$.
The space encoder operates on every frame and we obtain a sequence of features $[x^m_i]_{i=1}^{T}$.
Before feeding them to the time encoder, which is a transformer encoder, a learnable class token $\texttt{cls}$ is prepended to the sequence.
Its representation at the last layer of the encoder serves as the final representation used by the classification layer~\cite{devlin2018bert, dosovitskiy2020image, arnab2021vivit}.
\eqnsm{time_encoder_input}{\mathcal{X}_{(0)}^m = [\texttt{cls}^m, {x}_{1}^m, ... , {x}_{T}^m] + \texttt{pos} \,,}
where, $\texttt{pos}$ is a fixed absolute embedding which represent the temporal positions~\footnote{It is added to the tokens to maintain positional information since the subsequent self-attention operations in the transformers are permutation-invariant.}.
A sequence of tokens is then fed to a transformer encoder with $L$ layers (we set $L=2$), where each layer $l$ is composed of Multi-Head Self Attention (MHSA), Layer Normalization (LN), and Feed Forward Network (FFN)~\footnote{We leave an original paper as a reference~\cite{vaswani2017attention} for further details on transformer building blocks.}: 
\eqnsm{transformer1}{\mathcal{Y}_{(l)}^m = \textrm{MHSA}(\textrm{LN}(\mathcal{X}_{(l)}^m)) + \mathcal{X}_{(l)}^m \,,}
\eqnsm{transformer2}{\mathcal{X}_{(l+1)}^m = \textrm{FFN}(\textrm{LN}(\mathcal{Y}_{(l)}^m)) + \mathcal{Y}_{(l)}^m \,.}
The output at the final layer $\mathcal{X}_{(L)}^m \in \mathbb{R}^{(T+1) \times C}$ holds the same dimension as that of the input $\mathcal{X}_{(0)}^m$.
At last, we use the final classification token $\texttt{cls}_{L}^m \in \mathbb{R}^{C}$ to classify the action classes with the fully-connected (FC) layer.

\input{tabs/tips_and_tricks}

\input{tabs/fusion}

\begin{figure*}[t!]
    \centering
    \resizebox{\linewidth}{!}{
        \includegraphics[height=0.2\textheight]{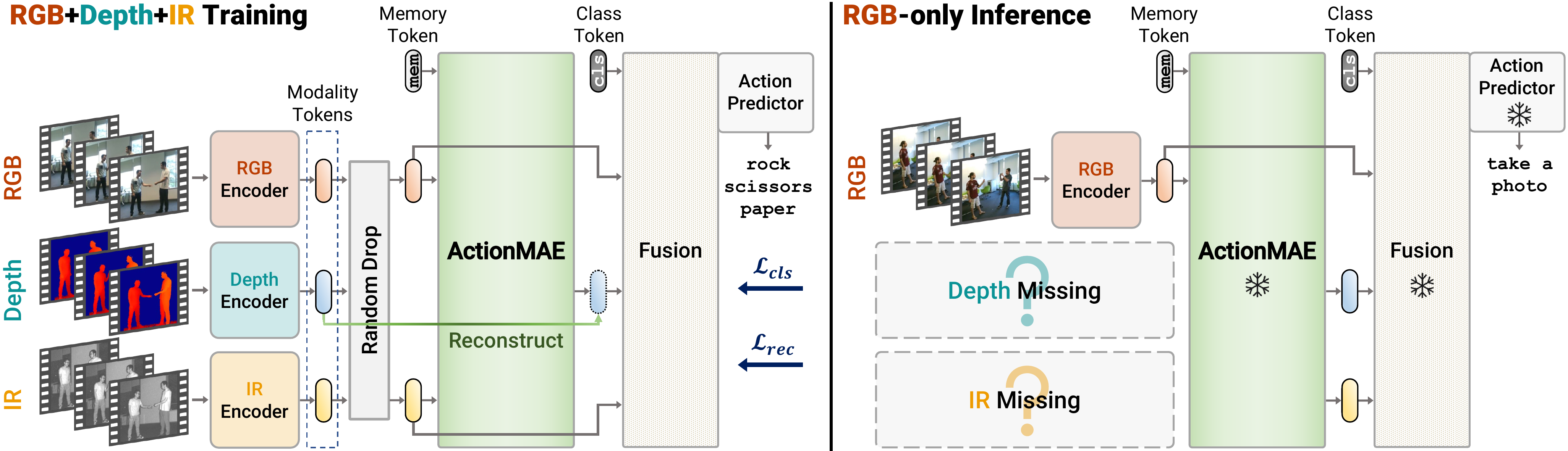}
    }
    \vspace{\abovefigcapmargin}
    \vspace{-3mm}
    \caption{
        {\ours framework.}
        \textbf{(i)} \textit{Multi-modal training}: We first obtain modality-specific features via their respective encoding stream.
        Then, we randomly drop $K$ (here, $K=1$) input tokens from $N$ (=3) modality tokens, where $K$ ($0 \leq K < N$) varies at every iteration. \ours learns to reconstruct missing modality tokens using the remaining $N-K$ modality tokens.
        During autoencoding, the memory token, which is also provided as an input, serves as global memory across varying inputs.
        The original and reconstructed tokens are reorganized as an input sequence to the fusion unit.
        The succeeding processes follow the standard fusion-and-predict procedure.
        The overall training objective is defined in~\Cref{eq:loss}.
        \textbf{(ii)} \textit{Missing modality inference}: We drop $K$ (=2) tokens (here, \depth, \ir) and predict the action class by reconstructing $K$ tokens with remaining $N-K$ (=1) token (\rgb).
        We note that \ours is agnostic to the types or the number of missing modalities.
    }
    \label{fig:actionmae}
    \vspace{\belowfigcapmargin}
\end{figure*}

\vspace{\subsecmargin}\subsection{Tips and Tricks for Training}
In our preliminary experiments, we noticed that our models are prone to overfit, highlighting the necessity for a bag of tricks to train our models effectively.
In order to see the effect of augmentations and regularizations, we ablate several strategies in~\Cref{tab:tips_and_tricks}.
By default, we employ center crop and horizontal flip for all modality data, and progressively add more strategies:
Imagenet initialization on space encoder is highly beneficial, implying that it eases the subsequent optimization of time encoder.
It has relatively large impact in the missing modality case compared to the full modality case, \ie, the strong inductive bias provided by ImageNet initialization can have a particularly positive effect in the absence of depth information.
Instead of the vanilla center crop, we center crop the larger region ($\times$1.2) and then random crop ($\times$1) within it.
Instead of uniformly sampling frames from video, we randomly sample frames and arrange them in temporal order.
Color jitter increases divergence of RGB distribution, which has a beneficial effect.
label smoothing~\cite{szegedy2016rethinking} adjusts the label distribution of the ground truth to encourage the model to produce less confident predictions.
Generally, regularizing the model or increasing the data diversity were beneficial in both full and missing modality cases.
However, some methods were detrimental in full modality case:
Color jitter on depth map rather harms the consistency of 3D structural information.
Dropout~\cite{srivastava2014dropout} makes optimization rather difficult.
In missing modality case, Dropout did not have a detrimental effect, but it also did not help regularize the model.
Overall, we achieve a substantial improvement of 7.4\%p (7.7\%p for missing modality case) on NTU RGB+D 60.
Note that no spatial or temporal augmentations are used in testing.

\vspace{\subsecmargin}\subsection{Fusion}
In~\Cref{tab:fusion}, we examine exemplary fusion methods in terms of multi-modal inference and missing modality inference.
We take three simple yet effective fusion methods that are commonly used in various multi-modal models: summation, concatenation, and transformer.
\begin{itemize}[nosep]
\item \textbf{Sum.} Each modality passes through its respective FC layer and then is all summed up:
$\sum\limits_{m} \textrm{FC}^{m}(\texttt{cls}_{L}^m) \,.$
\item \textbf{Concat.} Modalities are channel-wise concatenated ($\mathlarger{\mathlarger{\parallel}}$) and passed through a single FC layer: $\textrm{FC}(\underset{m}{\mathlarger{\mathlarger{\parallel}}} \texttt{cls}_{L}^m) \,.$~\footnote{In the case of concatenation, uni-modal predictions are approximated by splitting the weight matrix of FC layer into sub-matrices and dividing the bias by the number of modalities.}
\item \textbf{Transformer.} Extra class token is prepended to a sequence of modality tokens $[\texttt{cls}, \texttt{cls}_{L}^R, \texttt{cls}_{L}^D, \texttt{cls}_{L}^I]$, which is then passed through the transformer~(\cref{eq:transformer1,eq:transformer2}). The class token of the final transformer layer is subsequently passed through the FC layer.
\end{itemize}

The test accuracies with complete modalities (rows makred in \colorbox{Gray}{gray}) are not much different between three fusion methods.
Among them, the transformer-based fusion method is the most inferior (lags behind sum by 0.7\%p).
To our surprise, however, we observe the opposite tendency in a missing modality situation.
The transformer-based fusion is the most robust to the missing modalities: it reduces the mean accuracy discrepancy ($|{\bf \bar{\rm \triangle}}|$) by 13.8$\%$p compared to sum fusion.
Nevertheless, we notice the bias in all three fusion methods.
They specifically works well with a certain modality (\eg, depth), but poorly with other modalities (\eg, RGB, IR) under the situation of missing modality.

%% file: tabs/architecture.tex
\begin{table}[h!]\vspace{-.3em}
\tablestyle{1pt}{1.05}
\begin{tabular}{x{60}x{60}x{30}|x{40}x{40}}
\multirow{2}{*}{Space encoder$^*$}&\multirow{2}{*}{Time encoder}&\multirow{2}{*}{Fusion}&\multicolumn{2}{c}{Train $\rightarrow$ Test} \\
& & & {\R}{\D} $\rightarrow$ {\R}{\D} & {\R}{\D} $\rightarrow$ {\R} \\ 
\shline
\multicolumn{2}{c}{R(2+1)D}                 & \multirow{5}{*}{Sum}  & 89.2\%        & 41.2\% \\ 
ViT-tiny/16$^{\dagger}$     & Transformer   &                       & 81.4\%        & 35.7\% \\ 
ViT-small/16$^{\dagger}$    & Transformer   &                       & 83.0\%        & 36.3\% \\ 
ResNet18                    & Transformer   &                       & 91.9\%        & 41.9\% \\ 
\rowcolor{Gray}
ResNet34                    & Transformer   &                       & \textbf{92.6\%}        & \textbf{42.5\%} \\ 
\multicolumn{5}{c}{*Space encoders are Imagenet initialized.} \\
\multicolumn{5}{c}{$\dagger$~Re-implementations from~\cite{rw2019timm}.} \\
\end{tabular}
\vspace{\abovetabcapmargin}
\caption{
{ResNet34 + Transformer achieves the best accuracy in both full ({\R}{\D} $\rightarrow$ {\R}{\D}) and missing modality ({\R}{\D} $\rightarrow$ {\R}) cases.}
Experiments are conducted on NTU-RGB+D 60.
}
\vspace{\belowtabcapmargin}
\label{tab:architecture}
\end{table}

%% file: tabs/tips_and_tricks.tex
\begin{table}[t!]
\tablestyle{2pt}{1.1}
\begin{tabular}{y{125}|z{50}z{50}}
\multirow{2}{*}{Variant}&\multicolumn{2}{c}{Train$\rightarrow$Test} \\ 
& {\R}{\D}{\I} $\rightarrow$ {\R}{\D}{\I} & {\R}{\D} $\rightarrow$ {\R} \\
\shline
Center crop, horizontal flip    & 85.9 & 34.8 \\
+ Imagenet initialization       & 89.2~(\pacc{$+$3.3}) & 39.5~(\pacc{$+$4.7}) \\
+ Shifted center crop           & 90.5~(\pacc{$+$1.3}) & 40.3~(\pacc{$+$0.8}) \\
+ Temporal random sampling      & 91.0~(\pacc{$+$0.5}) & 41.4~(\pacc{$+$1.1}) \\
+ Color jitter ({\rgb})            & 92.2~(\pacc{$+$1.2}) & 42.0~(\pacc{$+$0.6}) \\
\rowcolor{Gray}
+ Label smoothing ($\alpha=0.1$)         & \textbf{93.3}~(\pacc{$+$1.1}) & \textbf{42.5}~(\pacc{$+$0.5}) \\
\hline
- Color jitter ({\depth})          & 91.5~(\macc{$-$1.8}) & - \\
- Dropout ($p=0.1$)             & 92.6~(\macc{$-$0.7}) & 42.5 \\
\end{tabular}
\vspace{\abovetabcapmargin}
\caption{
{Progressively adding regularizations helps much in both full and missing modality cases.}
However, applying color jittering on depth maps and using dropout rather reduced the performance in full modality case.
Experiments are conducted on NTU RGB+D 60 with the sum fusion model.
}
\vspace{\belowtabcapmargin}
\label{tab:tips_and_tricks}
\end{table}

%% file: tabs/fusion.tex
\begin{table}[t!]
\tablestyle{2pt}{1.1}
\begin{tabular}{y{48}|x{46}z{58}z{36}|x{30}}
Fusion & Train modal & Test modal\up & $\triangle$(\%p)\up & $|{\bf \bar{\rm \triangle}}|$\down \\
\shline
\multirow{7}{*}{Sum}   & \multirow{7}{*}{{\R}+{\D}+{\I}} & \cellcolor{Gray} {\R}+{\D}+{\I} (\textbf{93.3\%})  & \cellcolor{Gray} - & \multirow{7}{*}{\pacc{34.8}} \\
    &    & {\R}+{\D} (89.0\%)      &~\macc{$-$4.3}  & \\
    &    & {\R}+{\I} (59.4\%)      &~\macc{$-$33.9} & \\
    &    & {\D}+{\I} (89.2\%)      &~\macc{$-$4.1}  & \\
    &    & {\R}   (15.6\%)      &~\macc{$-$77.7} & \\
    &    & {\D}   (85.9\%)      &~\macc{$-$7.4}  & \\
    &    & {\I}   (12.1\%)      &~\macc{$-$81.2} & \\
\hline
\multirow{7}{*}{Concat}   & \multirow{7}{*}{{\R}+{\D}+{\I}} & \cellcolor{Gray} {\R}+{\D}+{\I} (\textbf{93.2\%})  & \cellcolor{Gray} -  & \multirow{7}{*}{\pacc{33.5}} \\
    &    & {\R}+{\D} (91.4\%)      &~\macc{$-$1.8}  & \\
    &    & {\R}+{\I} (75.4\%)      &~\macc{$-$17.8} & \\
    &    & {\D}+{\I} (18.6\%)      &~\macc{$-$74.6} & \\
    &    & {\R}   (53.1\%)      &~\macc{$-$40.1} & \\
    &    & {\D}   (88.2\%)      &~\macc{$-$5.0}  & \\
    &    & {\I}   (31.7\%)      &~\macc{$-$61.5} & \\
\hline
\multirow{7}{*}{Transformer}   & \multirow{7}{*}{{\R}+{\D}+{\I}} & \cellcolor{Gray} {\R}+{\D}+{\I} (\textbf{92.6\%})  & \cellcolor{Gray} - & \multirow{7}{*}{\pacc{21.0}} \\
    &    & {\R}+{\D} (91.6\%)      &~\macc{$-$1.0}  & \\
    &    & {\R}+{\I} (66.6\%)      &~\macc{$-$26.0} & \\
    &    & {\D}+{\I} (90.4\%)      &~\macc{$-$2.2}  & \\
    &    & {\R}   (56.6\%)      &~\macc{$-$36.0} & \\
    &    & {\D}   (88.9\%)      &~\macc{$-$3.7}  & \\
    &    & {\I}   (35.4\%)      &~\macc{$-$57.2} & \\
\end{tabular}
\vspace{\abovetabcapmargin}
\caption{
{Transformer is a missing modality robust fusion unit.}
Experiments are conducted on NTU RGB+D 60.
$\triangle$ is defined as accuracy discrepancy between missing modality and full modality cases ($\triangle = \textrm{Acc}_{miss} - \textrm{Acc}_{full}$).
While all three fusion methods show comparable performance when test modalities are complete, they show different aspects in missing modality inference. The rows marked in \colorbox{Gray}{gray} are complete modality cases.~{{\up}/{\down}: \rm The higher/lower the better.}
}
\vspace{\belowtabcapmargin}
\label{tab:fusion}
\end{table}

%% file: 03method.tex
\vspace{\secmargin}\section{Method: ActionMAE}
To address the missing modality scenario in action recognition problem, we propose \ours that learns missing modality predictive coding as illustrated in~\Cref{fig:actionmae}.

Our \ours is a simple modular autoencoder that reconstructs the missing modalities based on partial observation of the remaining modalities.
Like generic autoencoders, our solution is made up of an encoder that maps the observed modalities to a latent representation and a decoder that reconstructs the missing modalities from the latent representation.
The encoder and decoder adopted in our \ours are simple transformer encoders~\cite{vaswani2017attention}.
By design, it can be plugged into a typical multi-modal action classifier with a fusion unit.
We plug-and-play \ours on top of the model explored in preliminary experiments: ResNet34 + Transformer.
During training, the model equipped with \ours simultaneously learns missing modality information and predicts the correct action class based on these.

\vspace{\subsecmargin}\subsection{Missing Modality Predictive Coding}
We define a missing modality scenario as a case in which $N$ modalities are accessible during training, but $K$ modalities are missing during inference, \ie, only $N-K$ modalities are accessible, where $1 \leq K < N$.
We use three modalities in practice, RGB, Depth, and IR (as such, $N=3$), but the number or type of modalities are not limited to these.
From $N$ modality tokens $[z^m]_{m=1}^{N}$ (equivalent to $[\texttt{cls}_L^m]_{m=1}^{N}$ of~\cref{eq:transformer2}), we randomly drop $K$ tokens, \ie, only leaving $N-K$ tokens, where $0 \leq K < N$ (if $K=0$, we do not drop any of modality tokens). 
In addition to remaining $N-K$ tokens, the \ours encoder takes extra learnable token, \ie, memory token, that acts as global memory across varying inputs.
It is worth noting that the input sequence to \ours encoder continues to vary with each iteration.
We first obtain a sequence of latent representations via the encoder.
Then, the $K$ dummy tokens are inserted into the sequence at the position where the modality tokens are missing, and the resulting sequence is fed to the decoder.
As the dummy tokens pass through the decoder, they are forced to mimic the dropped tokens.
The original modality tokens are reorganized with the reconstructed tokens and passed to the subsequent fusion unit and prediction layer in turns.
Notice that the memory token is used only for encoding-decoding process, not for prediction.

\vspace{\subsecmargin}\subsection{Training \& Inference}
Overall training of the model with \ours is governed by the two loss functions: standard classification loss (\ie, cross-entropy loss) $\mathcal{L}_{\rm cls}$ and reconstruction loss $\mathcal{L}_{\rm rec}$.
The decoder aims to convert dummy tokens into original modality tokens.
We define the indices of missing tokens as $j_{1}, ... , j_{K}$. 
The reconstruction loss computes the mean squared error (MSE) between the reconstructed $[\hat{z}^{j_{1}}, ... , \hat{z}^{j_{K}}]$ and missing tokens $[z^{j_{1}}, ... , z^{j_{K}}]$ in the feature space:
\eqnsm{recon_loss}{\mathcal{L}_{rec} = \frac{1}{K} \sum_{i=1}^{K} (z^{j_{i}}-\hat{z}^{j_{i}})^2\,.}
Note that only memory token and dummy tokens are affected by the reconstruction loss (\ie, reconstruction loss is not computed for the remaining tokens) during training.

Our training objective is jointly optimized upon the two loss functions:
\eqnsm{loss}{\mathcal{L} = \lambda_{\rm cls}\mathcal{L}_{\rm cls} + \lambda_{\rm rec}\mathcal{L}_{rec}\,.}
In our experiments, we empirically find that balancing hyperparameters $\lambda_{\rm cls}=\lambda_{\rm rec}=1$ works well.

At inference time, $K$ modalities are dropped from the input stage, and remaining $N-K$ modalities are encoded by their respective modality encoders. 
\ours takes only the memory token and $N-K$ encoded modality tokens, and reconstructs the missing tokens.
We progressively drop the modalities to see whether \ours is robust under all circumstances, regardless of which modality or how many modalities are lost.
Experiments demonstrate that our \ours is quite robust to any cases of modality missing.

%% file: 04experiments.tex
\vspace{\secmargin}\section{Experiments}

\input{tabs/actionmae_missing}

\input{tabs/benchmarks}

\vspace{\subsecmargin}\subsection{Implementation}
\vspace{\paramargin}\paragraph{Inputs.}
The input of each modality network is a 16 frame clip of the corresponding modality, and their input size is 224$\times$224 pixels.
In the spatial domain, random horizontal flip and shifted center crop are performed on all modality inputs.
RGB frames are applied with color jittering, and the 1-channel depth frames are converted into 3-channels based on a JET color map with 256 scales.
In the temporal domain, we randomly sampled 16 frames from the raw video, and arrange them in temporal order.
At test time, each clip consists of 16 frames that are uniformly sampled, center cropped, and no data augmentation techniques are applied.

\vspace{\paramargin}\paragraph{Settings.}
We used AdamW optimizer~\cite{loshchilov2017decoupled} with an initial learning rate of $10^{-4}$ and weight decay of $10^{-4}$ for a batch size of 32.
The learning rate is linearly decayed by a factor of 10 every 30 epochs.
We used cross-entropy loss with label smoothing of factor 0.1~\cite{szegedy2016rethinking}.
All space encoders are initialized with imagenet pre-trained weights~\cite{deng2009imagenet}.
\ours features a symmetrical design with two encoder and decoder layers.

\vspace{\subsecmargin}\subsection{Datasets}
\noindent\textbf{NTU RGB+D 60 (NTU60)}~\cite{shahroudy2016ntu} includes 60 classes and 56,880 video samples.
The actions are in three categories: daily actions, mutual actions, and medical conditions.
The dataset is captured by three Kinect V2 cameras concurrently, which contains RGB videos (spatial size: $480\times270$), depth map sequences ($512\times424$), infrared ($512\times424$), and 3D skeletal videos ($25$ joints).

\vspace{2mm}\noindent\textbf{NTU RGB+D 120 (NTU120)}~\cite{liu2019ntu} extends NTU RGB+D 60 by adding an additional 60 classes and 57,600 video samples, bringing the total to 120 classes and 114,480 samples.
It comprises 106 subjects across 155 different viewpoints and 96 background scenes.
Following the convention~\cite{liu2019ntu}, we evaluate our model using the cross-subject protocol in both NTU60 and NTU 120.

\vspace{2mm}\noindent\textbf{NW-UCLA}~\cite{wang2014cross} contains 10 action classes and 1,475 video samples collected from 10 different subjects.
RGB ($640\times480$), depth ($320\times240$), and skeletal videos ($20$ joints) are captured simultaneously from 3 different viewpoints with Kinetic V1.
We followed the cross-view protocol suggested in~\cite{wang2014cross}, using two views (V1, V2) for training and the remaining (V3) for testing.

\vspace{2mm}\noindent\textbf{UWA3DII}~\cite{rahmani2016histogram} consists of 30 classes and 1,070 videos: RGB ($320\times240$), depth ($320\times240$), and skeleton ($15$ joints).
The actions are performed by 10 subjects in 4 different views: front (V1), left side (V2), right side (V3), and top view (V4).
We used the last two views (V3, V4) for training and the remaining views (V1, V2) for testing.

\vspace{\subsecmargin}\subsection{ActionMAE}
In \Cref{tab:actionmae_missing}, we examined the effect of \ours by comparing the model with and without \ours.
In preliminary experiments, we observed that complete modality performance was rather lower when using transformer than much simpler methods such as sum or concatenation.
We hypothesize that this is due to the lack of inductive bias in transformer, which makes it susceptible to overfitting.
When we attach \ours on a transformer-fusion model, the random drop operation yields nontrivial self-supervisory task, which effectively regularizes the model.
This results in $0.4\%$p performance improvement under complete modality setting.
Generally, \ours shows good robustness to any combination of missing modalities.
More importantly, the missing modality performances, which were particularly poor in accuracy, are significantly improved upon vanilla transformer baseline (\eg, IR: $35.4\% \rightarrow 81.7\%$).
This implies that \ours also has a debiasing effect on the dominant modality (\eg, depth).
Overall, \ours reduces the mean accuracy discrepancy by $\sim 2.5\times$ compared to the transformer-fusion baseline, and by $\sim 4\times$ compared to the sum-fusion baseline.

\vspace{\subsecmargin}\subsection{Comparative Study}
In~\Cref{tab:benchmarks}, we benchmark our approach against competitive baselines on multiple action recognition datasets.

\vspace{\paramargin}\paragraph{\textit{Vs.} uni-/multi-modal baselines.}
We begin by comparing \ours with transformer-fusion baselines.
With complete modalities, \ours mostly improves performance upon multi-modal baselines as randomized drop of modalities during training gives a regularization effect on the overall network analogous to dropout~\cite{srivastava2014dropout}.
While our model is also fairly robust to missing modalities in both R+D+I and R+D cases, uni-modal models outperform \ours when tested with a single modality.
However, in the missing modality situation evaluated with R+D, our model (92.6\%) slightly improves accuracy compared to the multi-modal baseline (92.3\%).

\vspace{\paramargin}\paragraph{\textit{Vs.} complete modality counterparts.}
We compare our approach with recent action recognition methods.
These methods do not drop any modality at the test time (\ie, complete modality inference), yet our method significantly outperforms all listed methods even in missing modality scenarios.

\input{tabs/ablations}

\vspace{\paramargin}\paragraph{\textit{Vs.} missing modality counterparts.}
Next, we compare our approach with missing modality counterparts.
Generally, as more modalities are used in the training stage, they show better performance in missing modality situations.
In both R+D+I and R+D cases, the accuracy of \ours decreases as more modalities are dropped, albeit shows competitive performance.
When we follow the most common setup in the literature (train: R+D, test: R or D), \ours sets new state-of-the-art results.

\vspace{\paramargin}\paragraph{\ours (R+D+I \vs R+D).}
In the case of R+D+I, if just one modality is missing at test time (\eg, R+D), there is a minimal loss of accuracy ($-$0.4\%p).
However, in the absence of two modalities (\eg, R), mapping from latent representation becomes more complex as the number of values to be restored increases, thus drops more ($-$9.6\%p).
On the other hand, even evaluated with the same modality (\eg, D), R+D performs better than R+D+I since the mapping is simpler.

\vspace{\subsecmargin}\subsection{Ablative Study}
From~\Cref{tab:ablations}, we observed several intriguing properties of \ours by ablating it.

\vspace{\paramargin}\paragraph{Memory token.}
Memory token contributes significantly to performance improvement, especially for the missing modality cases: 9.7\%p gain for R-only case and 7.3\%p gain for D-only case.
We understand this as memory token learns meaningful correlation between varying inputs and models pair-wise relationships between itself and modality tokens.

\vspace{\paramargin}\paragraph{Loss hyperparameters.}
We investigated a good balance between reconstruction $\mathcal{L}_{rec}$ and classification loss $\mathcal{L}_{cls}$.
From an exploration-exploitation perspective, when the overall objective is leaned to classification (\ie, $\mathcal{L}_{rec}:\mathcal{L}_{cls}=1:10$), \ours underutilizes the features extracted from the space-time encoder, resulting in poor `exploitation' of learned information.
On the other hand, if the objective is leaned to reconstruction (\ie, 10:1), \ours loses the chance of `exploration' to generate more class-discriminative features.
We found a ratio of 1:1 properly balances the two losses.

\vspace{\paramargin}\paragraph{Position embedding.}
We observed a significant performance improvements with position embedding (\texttt{pos}).
From the results, we hypothesize that position embedding serves as a reference for determining which modalities needs to be reconstructed in \ours, \ie, position embedding provides guidance on what to reconstruct.

\vspace{\paramargin}\paragraph{Pre-training.}
We examined whether pre-training the space-time encoder would ease the optimization or rather negatively impact the training of \ours.
Here, PT (S) indicates space encoder pre-training, and PT (S+T) indicates pre-training both space and time encoders.
Contrary to our expectations, pre-training any encoders ahead of \ours diminishes accuracy.
Even, the performance degraded when more layers were pre-trained.
This suggests that \ours can be effectively learned through self-supervisory reconstruction of diverse inputs.
However, if the space-time encoder is pre-trained, the input will remain almost constant; hence, \ours will likely to find a trivial solution.
Therefore, we only performed imagenet initialization on the space encoder and trained all networks from scratch.


%% file: tabs/actionmae_missing.tex
\begin{table}[t!]
\tablestyle{2pt}{1.1}
\begin{tabular}{y{54}|x{44}z{56}z{36}|x{28}}
Model & Train modal & Test modal\up & $\triangle$(\%p)\up & $|{\bf \bar{\rm \triangle}}|$\down \\
\shline
Sum & \multirow{3}{*}{{\R}+{\D}+{\I}}& \cellcolor{Gray} {\R}+{\D}+{\I} (93.3\%)  & \cellcolor{Gray} - &~\pacc{34.8}\\
Concat &                    & \cellcolor{Gray} {\R}+{\D}+{\I} (93.2\%)  & \cellcolor{Gray} - &~\pacc{33.5}\\
Transformer &               & \cellcolor{Gray} {\R}+{\D}+{\I} (92.6\%)  & \cellcolor{Gray} - &~\pacc{21.0}\\
\hline
\multirow{7}{*}{\begin{tabular}[l]{@{}c@{}}Transformer\\ w/ \ours \end{tabular}}   & \multirow{7}{*}{{\R}+{\D}+{\I}} & \cellcolor{Gray} {\R}+{\D}+{\I} (93.0\%)  & \cellcolor{Gray} - & \multirow{7}{*}{\pacc{8.3}} \\
                                &                       & {\R}+{\D} (92.6\%)      &~\macc{$-$0.4}  & \\
                                &                       & {\R}+{\I} (87.6\%)      &~\macc{$-$5.4}  & \\
                                &                       & {\D}+{\I} (91.9\%)      &~\macc{$-$1.1}  & \\
                                &                       & {\R}   (83.4\%)      &~\macc{$-$9.6} & \\
                                &                       & {\D}   (89.7\%)      &~\macc{$-$3.3}  & \\
                                &                       & {\I}   (81.7\%)      &~\macc{$-$11.3} & \\
\end{tabular}
\vspace{\abovetabcapmargin}
\caption{
{\ours is a missing modality robust learner.}
Experiments are conducted on NTU RGB+D 60.
The rows marked in \colorbox{Gray}{gray} are complete modality cases.~{{\up}/{\down}: \rm The higher/lower the better.}
}
\vspace{\belowtabcapmargin}
\label{tab:actionmae_missing}
\end{table}

%% file: tabs/benchmarks.tex
\begin{table*}[t!]
\tablestyle{2pt}{1.1}
\resizebox{\linewidth}{!}{
\begin{tabular}{y{10}y{160}x{50}x{50}x{40}x{42}x{42}x{42}x{42}}
\shline
 & Model & Train modal$^{*}$ & Test modal & Universal$^{\dagger}$ & NTU60 & NTU120 & NWUCLA & UWA3D \\
\shline
\multirow{5}{*}{\rot{\textit{Baseline}}}    & \multirow{3}{*}{Uni-modal (/w Transformer-fusion)}   & {\R}         & {\R}         & \xmark & 86.6\% & 84.2\% & 85.0\% & 73.1\% \\
                                            &                                                      & {\D}         & {\D}         & \xmark & 92.0\% & 88.2\% & 92.7\% & 81.2\% \\
                                            &                                                      & {\I}         & {\I}         & \xmark & 85.5\% & 85.4\% &    -   &    -   \\
                                            & \multirow{2}{*}{Multi-modal (/w Transformer-fusion)} & {\R}+{\D}       & {\R}+{\D}       & \xmark & 92.3\% & 91.7\% & 90.9\% & 82.5\% \\
                                            &                                                      & {\R}+{\D}+{\I}     & {\R}+{\D}+{\I}     & \xmark & 92.6\% & 92.1\% &    -   &    -   \\
\hline
\multirow{4}{*}{\rot{\textit{Complete}}}    & 3DFCNN~\cite{sanchez20223dfcnn}       & {\D}     & {\D}     & \xmark & 78.1\% &    -   & 83.6\% & 66.6\% \\
                                            & Ren \etal~\shortcite{ren2021multi}               & {\R}+{\D}   & {\R}+{\D}   & \xmark & 89.7\% &    -   &    -   &    -   \\
                                            & Deep Bilinear~\cite{hu2018deep}       & {\R}+{\D}+S & {\R}+{\D}+S & \xmark & 85.4\% &    -   &    -   &    -   \\
                                            & DMCL-complete~\cite{garcia2019dmcl}   & {\R}+{\D}+F & {\R}+{\D}+F & \xmark & 87.3\% & 89.7\% & 93.9\% & 89.8\% \\
\hline
\multirow{7}{*}{\rot{\textit{Missing}}}     & Hoffman \etal~\shortcite{hoffman2016learning}                 & {\R}+{\D}                       & {\R}    & \xmark &    -              &    -              & {83.2\%}     & {66.7\%}     \\
                                            & Garcia \etal~\shortcite{garcia2018modality}                  & {\R}+{\D}                       & {\R}    & \xmark & {73.1\%}     &    -              & {86.7\%}     & {73.2\%}     \\
                                            & ADMD~\cite{garcia2019learning}                & {\R}+{\D}                       & {\R}    & \xmark & {73.4\%}     &    -              & {93.6\%}     & {78.4\%}     \\
                                            \cmidrule(l){2-9}
                                            & \multirow{2}{*}{Luo \etal~\shortcite{luo2018graph}}       &\multirow{2}{*}{{\R}+{\D}+F+S}& {\R}    & \xmark & {89.5\%}     &    -              &    -              &    -              \\
                                            &                                               &                           & {\D}    & \xmark & {87.5\%}   &    -              &    -              &    -              \\
                                            \cmidrule(l){2-9}
                                            & \multirow{2}{*}{DMCL~\cite{garcia2019dmcl}}   & \multirow{2}{*}{{\R}+{\D}+F}    & {\R}    & \xmark & {83.6\%}     & {84.3\%}     & {93.6\%}     & {78.4\%}     \\
                                            &                                               &                           & {\D}    & \xmark & {80.6\%}   & {82.2\%}   & {83.3\%}   & {81.9\%}   \\
\hline
\multirow{10}{*}{\rot{\textit{Ours}}}       & \multirow{7}{*}{\ours ({\R}+{\D}+{\I})}    & \multirow{7}{*}{{\R}+{\D}+{\I}}    & {\R}+{\D}+{\I} & \cmark & \cellcolor{Gray} 93.0\%  & \cellcolor{Gray} 92.3\%   &    -                      &    -                      \\
                                            &                                   &                           & {\R}+{\D}   & \cmark & 92.6\%                   & 91.7\%                    &    -                      &    -                      \\
                                            &                                   &                           & {\R}+{\I}   & \cmark & 87.6\%                   & 84.8\%                    &    -                      &    -                      \\
                                            &                                   &                           & {\D}+{\I}   & \cmark & 91.9\%                   & 91.2\%                    &    -                      &    -                      \\
                                            &                                   &                           & {\R}     & \cmark & {83.4\%}            & {83.1\%}             &    -                      &    -                      \\
                                            &                                   &                           & {\D}     & \cmark & {90.1\%}          & {86.0\%}           &    -                      &    -                      \\
                                            &                                   &                           & {\I}     & \cmark & {81.7\%}             & {81.6\%}              &    -                      &    -                      \\
                                            \cmidrule(l){2-9}
                                            & \multirow{3}{*}{\ours ({\R}+{\D})}      & \multirow{3}{*}{{\R}+{\D}}      & {\R}+{\D}   & \cmark & \cellcolor{Gray} 92.5\%  & \cellcolor{Gray} 91.5\%   & \cellcolor{Gray} 91.0\%   & \cellcolor{Gray} 79.8\%   \\
                                            &                                   &                           & {\R}     & \cmark & {84.5\%}            & {84.7\%}             & {84.2\%}             & {70.4\%}             \\
                                            &                                   &                           & {\D}     & \cmark & {90.5\%}          & {87.0\%}           & {88.2\%}           & {77.6\%}           \\
\multicolumn{9}{c}{*\textbf{{\R}}: RGB, \textbf{{\D}}: depth, \textbf{{\I}}: infrared, \textbf{F}: optical flow, \textbf{S}: skeleton. $\dagger$ preserves accuracy no matter what kind and how many modalities are missing.}
\end{tabular}
}
\vspace{\abovetabcapmargin}
\caption{
{Comparative study} on multiple action recognition benchmarks: NTU60~\cite{shahroudy2016ntu}, NTU120~\cite{liu2019ntu}, NWUCLA~\cite{wang2014cross}, and UWA3D~\cite{rahmani2016histogram}.
Here, we report top-1 accuracy.
The rows marked in \colorbox{Gray}{gray} are complete modality cases of ours.
By design, our \ours is robust to missing modalities of any type and number.
}
\vspace{\belowtabcapmargin}
\vspace{-1mm}
\label{tab:benchmarks}
\end{table*}

%% file: tabs/ablations.tex
\begin{table*}[t!]
\centering
\subfloat[
\textbf{Memory token.}
\label{tab:memory_token}
]{
\begin{minipage}{0.22\linewidth}{\begin{center}
\tablestyle{3pt}{1.05}
\begin{tabular}{lx{15}x{15}x{15}}
case                & {\R}{\D}                   & {\R}             & {\D}           \\
\shline
w/o \texttt{mem}    & 87.3                 & 74.8                 & 83.2                 \\
w/ \texttt{mem}     & \cellcolor{Gray}92.5 & \cellcolor{Gray}84.5 & \cellcolor{Gray}90.5 \\
\multicolumn{4}{c}{~}\\
\end{tabular}
\end{center}}\end{minipage}
\vspace{-1mm}
}
\hspace{1em}
\subfloat[
\textbf{Loss coefficients.}
 \label{tab:loss_hyperparameter}
]{
\begin{minipage}{0.22\linewidth}{\begin{center}
\tablestyle{3pt}{1.05}
\begin{tabular}{lx{15}x{15}x{15}}
$\mathcal{L}_{rec}$:$\mathcal{L}_{cls}$ & {\R}{\D}                       & {\R}             & {\D}           \\
\shline
1:10                                    & 91.9                     & 83.0                 & 90.0                 \\
1:1                                     & \cellcolor{Gray}92.5     & \cellcolor{Gray}84.5 & \cellcolor{Gray}90.5 \\
10:1                                    & 91.0                     & 81.6                 & 88.7                 \\
\end{tabular}
\end{center}}\end{minipage}
\vspace{-1mm}
}
\hspace{1em}
\subfloat[
\textbf{Position embedding.}
 \label{tab:position_embedding}
]{
\begin{minipage}{0.22\linewidth}{\begin{center}
\tablestyle{3pt}{1.05}
\begin{tabular}{lx{15}x{15}x{15}}
case                            & {\R}{\D}                   & {\R}             & {\D}           \\
\shline
w/o \texttt{pos}                & 87.2                 & 76.3                 & 84.7                 \\
w/ \texttt{pos}                 & \cellcolor{Gray}92.5 & \cellcolor{Gray}84.5 & \cellcolor{Gray}90.5 \\
\multicolumn{4}{c}{~}\\
\end{tabular}
\end{center}}\end{minipage}
\vspace{-1mm}
}
\hspace{1em}
\subfloat[
\textbf{Pre-training.}
 \label{tab:pretraining}
]{
\begin{minipage}{0.22\linewidth}{\begin{center}
\tablestyle{3pt}{1.05}
\begin{tabular}{lx{15}x{15}x{15}}
case     & {\R}{\D}                   & {\R}             & {\D}           \\
\shline
No PT    & \cellcolor{Gray}92.5 & \cellcolor{Gray}84.5 & \cellcolor{Gray}90.5 \\
PT (S)   & 89.1                 & 82.0                 & 86.4                 \\
PT (S+T) & 88.5                 & 81.9                 & 86.6                 \\
\end{tabular}
\end{center}}\end{minipage}
\vspace{-1mm}
}
\\
\vspace{\abovetabcapmargin}
\caption{
{Ablation experiments} on NTU RGB+D 60 with \ours ({\R}+{\D}).
The entries marked in \colorbox{Gray}{gray} are our settings.
\label{tab:ablations}
}
\vspace{\belowtabcapmargin}
\vspace{-2mm}
\end{table*}

%% file: 05related_work.tex
\vspace{\secmargin}\section{Related Work}
Multi-modality generally yields reliable results since different modalities provide complementary information~\cite{huang2021makes}.
For example, RGB provides rich appearance information, depth provides 3D geometrical structure, and IR is robust to illumination variation.
The missing modality setup assumes certain modalities available during training are unavailable at test time.
As the typical multi-modal models suffer in such scenarios, there was a large body of studies to address the missing modality problem~\cite{vapnik2009new,pechyony2010theory,vapnik2015learning,lopez2015unifying,ma2022multimodal,zhao2021missing,alayrac2020self,ma2021smil} 

In the context of action recognition~\cite{tran2015learning,wang2016temporal,carreira2017quo,feichtenhofer2019slowfast}, there are handful of studies towards this end~\cite{hoffman2016learning,luo2018graph,garcia2018modality,garcia2019learning,garcia2019dmcl,stroud2020d3d}, all of which follow transfer learning scheme via knowledge distillation~\cite{hinton2015distilling}.
We rather take inspiration from recent successes of masked autoencoders~\cite{devlin2018bert,he2022masked,feichtenhofer2022masked,bachmann2022multimae}, whose core idea is to remove a portion of the data and learn to predict the removed data, and show that masked autoencoding works better than the knowledge distillation approaches. In addition to accuracy, there are several advantages of \ours over the knowledge distillation approaches:
\textbf{(i)} A separate teacher network training, which is a prerequisite for knowledge distillation, is unnecessary. Therefore, the training can be completed in an end-to-end manner with only a single step.
\textbf{(ii)} We do not need method-specific architecture. \ours can be plugged into any type of pre-existing space-time encoders.
\textbf{(iii)} By design, \ours is universal to the type or the number of missing modalities.

%% file: 06discussion.tex
\vspace{\secmargin}\section{Discussion}
In this paper, we have answered the following questions regarding the missing modality scenarios in multi-modal action recognition problem:
\noindent\textbf{(i)} \textit{How to train a strong multi-modal model?}
We sought the good practices from three perspectives (see~\Cref{tab:architecture,tab:tips_and_tricks}): architecture, data augmentation, and regularization.
\noindent\textbf{(ii)} \textit{Which fusion should we choose for missing modality action recognition?} 
Of the three popular choices for fusion (\eg, sum, concatenation, transformer), transformer worked the best (see~\Cref{tab:fusion}), albeit still far from enough.
\noindent\textbf{(iii)} \textit{Is there more effective way to solve missing modality problem?} 
We showed that \ours is an effective strategy for missing modality action recognition via extensive experiments (see~\Cref{tab:actionmae_missing,tab:benchmarks,tab:ablations}).
In addition, it relieves bias against dominant modality, as well as effectively regularizes multi-modal model.
We hope that our findings will provide insights into broader missing modality scenarios.

\vspace{\paramargin}\paragraph{Limitations \& future work.}
There still remain unresolved issues:
\noindent\textbf{(i)} The uni-modal baselines perform better than the model equipped with \ours when evaluated with a single modality in missing modality situations.
\noindent\textbf{(ii)} Even with the same test modality (\eg, R or D), the model with fewer training modalities (R+D) outperforms the model with more training modalities (R+D+I).
We posit these challenges occur due to the difficulty of mapping from latent representation of \ours.
As the number of outputs exceeds the number of inputs, it yields a sub-optimal solution.
We mitigate this problem using memory token, yet more investigation is needed to find a more effective way.
Despite the limitations, our idea is simple and can be naturally extended to other modalities, such as vision, language, audio, \etc.
We leave this intriguing challenge to future work.



%% file: 07acknowledgement.tex
\section*{Acknowledgments}
This work was supported by the Agency For Defense Development by the Korean Government (UD190031RD).

%% file: 08appendix.tex
\appendix
\section*{Appendix}
\vspace{1mm}

\vspace{\secmargin}\section{Background: Transformer}
The standard transformer~\cite{vaswani2017attention} includes multi-head self-attention layers, residual connections, layer normalization, and feed forward networks.
As our model is built on the transformers, here we briefly explain how the multi-head self-attention, the core component of transformer, work internally. 

\vspace{2mm}\noindent\textbf{Self-Attention (SA)} is a popular yet strong mechanism for neural systems that enables inputs to interact with each another (``self") and find out which element they should pay more attention to (``attention").
The attention function is essentially a mapping between a query (Q) and a set of key-value (K-V) pairs.
The outputs are aggregations of these interactions and attention scores.
Formally, we compute a weighted sum over all values ${\bf V}$ for each element in an input sequence ${\bf X} \in \mathbb{R}^{N \times d}$.
The attention weights $A_{ij}$ are pair-wise similarity between two elements of the sequence and their respective query ${\bf Q}_i$ and key ${\bf K}_j$ representations.
\vspace{1mm}
\begin{equation}
    [{\bf Q, K, V}] = {\bf X}{{\bf W}_{QKV}} + {\bf p},
\end{equation}
\begin{equation}
    A = {\rm softmax}\left( \frac{{\rm\bf QK}^T}{\sqrt{D_h}} \right),
\end{equation}
\begin{equation}
    {\rm SA}({\bf X}) = A{\bf V},
\end{equation}
where \({{\bf W}_{QKV}} \in \mathbb{R}^{d \times 3d_h}\) is projection matrix that are used to generate different subspace representations of the query, key, value matrices, and \(A \in \mathbb{R}^{N \times N}\) is an attention matrix.
In practice, we add positional encoding \({\bf p} \in \mathbb{R}^{N \times d}\) to embedded sequence as the transformer is inherently permutation-invariant \wrt input sequence.

\vspace{2mm}\noindent\textbf{Multi-Head Self-Attention (MHSA)} is an extension of self-attention where $k$ self-attentions (\aka, "heads") are performed in parallel, and followed by a projection of their concatenated outputs.
Here, $d_h$ is generally set to $d/k$ to maintain the computation and the number of parameters constant when $k$ varies.
\begin{equation}
    {\rm MHSA({\bf x})} = [{\rm SA}_1({\bf x}); {\rm SA}_2({\bf x}); \cdots; {\rm SA}_k({\bf x});]{{\bf W}_{O}},
\end{equation}
where [;] denotes channel-wise concatenation and \({{\bf W}_{O}} \in \mathbb{R}^{k\cdot d_h \times d}\) denotes a projection matrix for the multi-head outputs.

\vspace{\secmargin}\section{More Details About \ours}
\begin{figure}[h!]
    \centering
        \includegraphics[width=\linewidth]{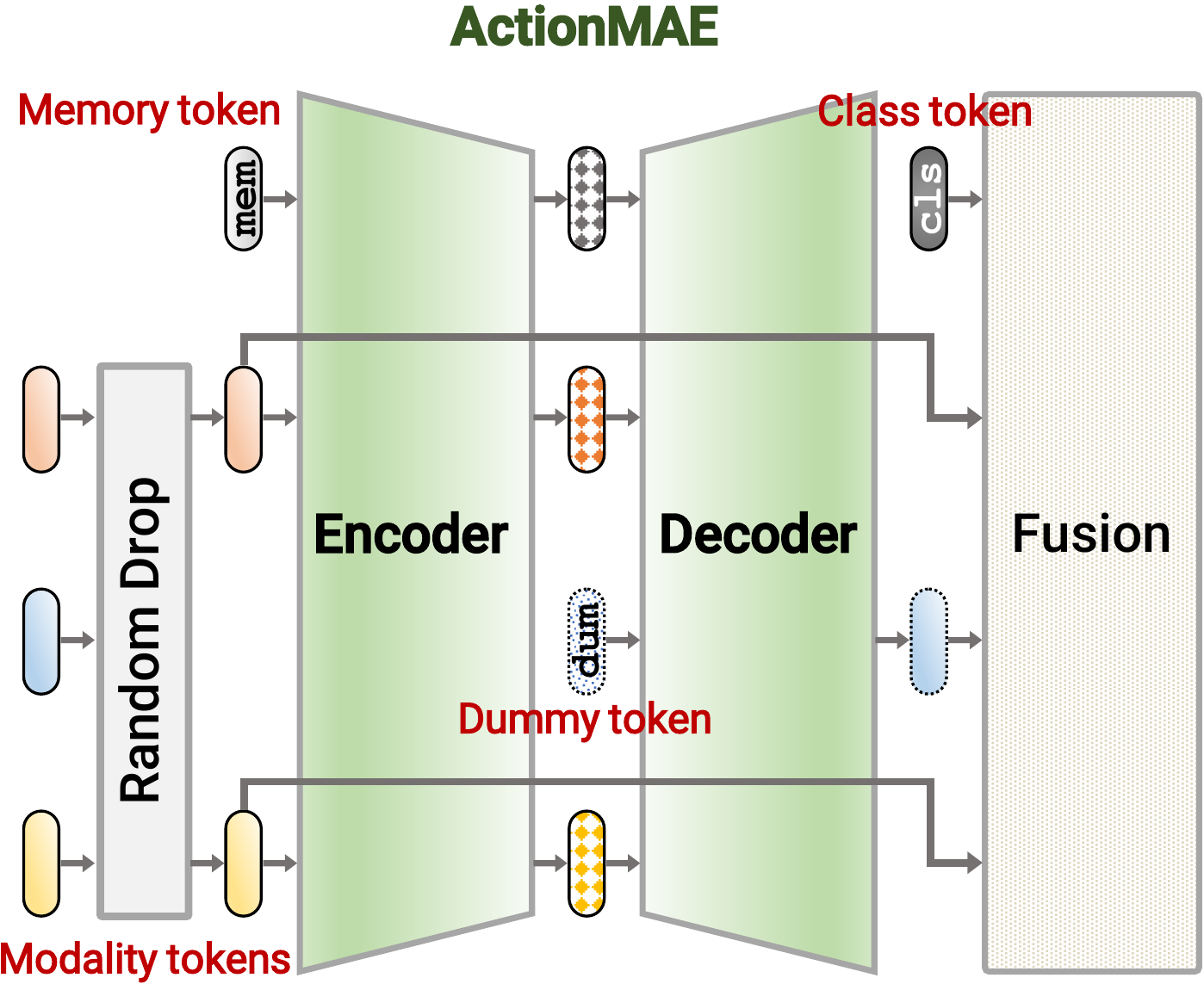}
    \vspace{\abovefigcapmargin}
    \caption{
        \textbf{High-level overview of \ours encoder-decoder.}
        \ours is a kind of autoencoder that uses an encoder-decoder architecture.
        This figure assumes a situation in which a blue modality token is dropped.
    }
    \label{fig:actionmae_enc_dec}
\end{figure}
\begin{figure}[t!]
    \centering
        \includegraphics[width=0.96\linewidth]{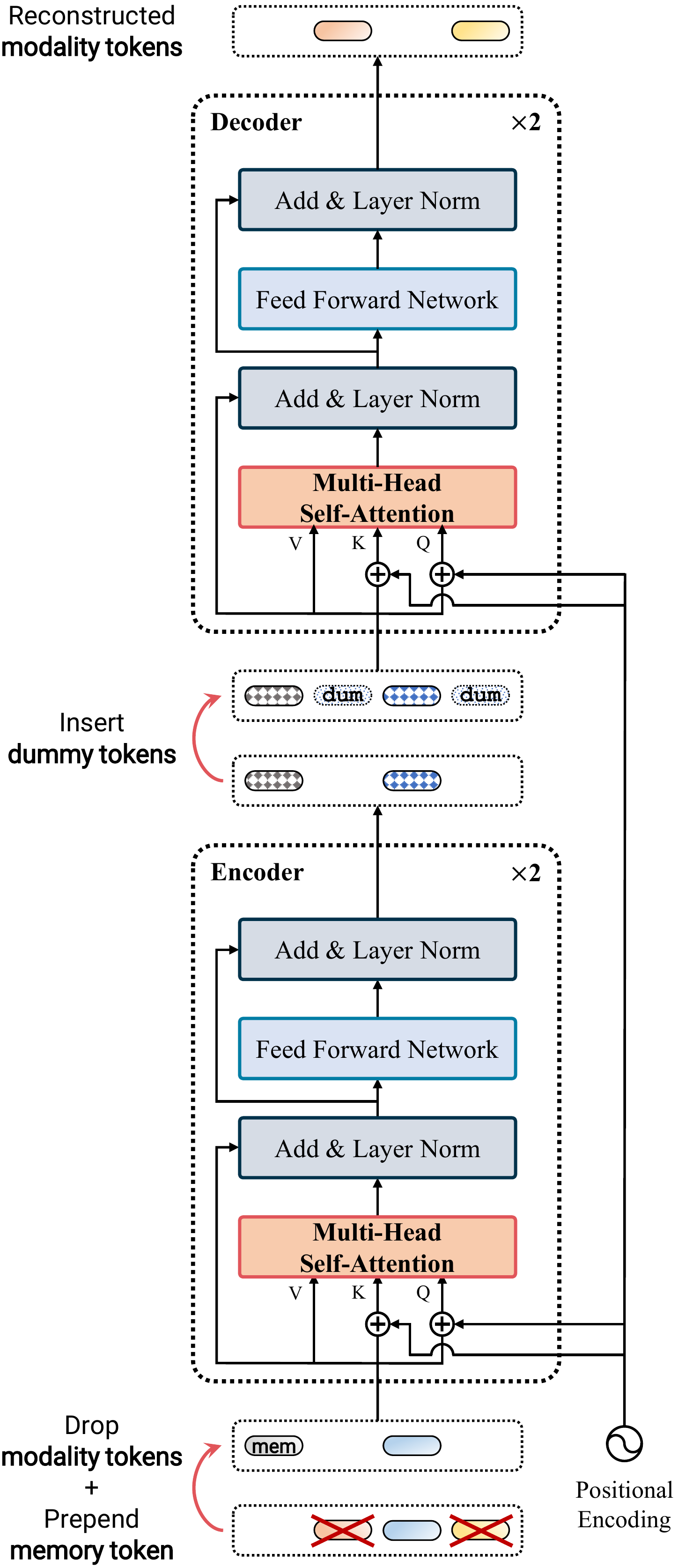}
    \vspace{\abovefigcapmargin}
    \caption{
        \textbf{Architectural details of \ours.}
        \ours is built with two layers of encoder and two layers of decoder
        \ours has symmetric design, with encoder and decoder made up of two transformer encoder layers each.
        The fixed absolute positional encoding~\cite{vaswani2017attention} is added at the beginning of the encoding or decoding operation.
        Here, we assume a situation in which the red and yellow modality tokens are dropped.
    }
    \label{fig:architecture}
    \vspace{\belowfigcapmargin}
\end{figure}
\Cref{fig:actionmae_enc_dec} shows a high-level overview of \ours encoder-decoder, and \Cref{fig:architecture} illustrates a more detailed architecture of \ours.
Given a sequence of $N$ modality tokens, $[z^1, \cdots, z^N]$, $K$ tokens are randomly removed such that $0 \leq K < N$, \ie, do not remove any token if $K=0$ and leave at least one token if $K=N-1$, resulting in $[z^{r_1}, \cdots, z^{r_{N-K}}]$, where $r_1, \cdots, r_{N-K}$ are indices of $N-K$ remaining tokens.
Note that $K$ varies with each iteration.
A memory token \texttt{mem} is prepended to the sequence and the resulting sequence, $[\texttt{mem}, z^{r_1}, \cdots, z^{r_{N-K}}]$, is fed to the \ours.
\ours, like all other autoencoders, comprises of an encoder and a decoder.
The encoder maps observed modalities to latent representations and the decoder reconstructs the representations of missing modalities from the latent representations.
During this autoencoding process, the memory token also serves as a global memory that learns pair-wise relationship between varying inputs.
Formally, the encoder maps an input sequence to latent representations, $[\texttt{mem}_e, z^{r_1}_e, \cdots, z^{r_{N-K}}_e]$.
We use learnable token, dubbed as dummy token \texttt{dum}, that learns to mimic the missing token.
$K$ dummy tokens, each recovers corresponding missing token, are inserted into the sequence at the position where the modality tokens were removed:
$[\texttt{mem}_e, z^{r_1}_e, \texttt{dum}^{j_1},  \cdots, \texttt{dum}^{j_K} , z^{r_{N-K}}_e]$ (the positional indices of missing tokens are defined as $j_1, \cdots, j_k$).
This sequence then passes through the \ours decoder, and generates missing tokens: $[\hat{z}^{j-1}, \cdots, \hat{z}^{K}]$.
Finally, a class token \texttt{cls} is prepended to reconstructed tokens, $[\texttt{cls}, \hat{z}^{j-1}, \cdots, \hat{z}^{K}]$, and the resulting sequence passes through the fusion unit followed by a prediction head.

In our framework, \ours learns to recover missing modality, which is then used for action classification along with the remaining modality.
By design, unlike previous KD-based methods that require retraining when missing modality is changed (when R+D $\rightarrow$ R is changed to R+D $\rightarrow$ D), \ours can handle all types of missing modalities (\ie, universal) with just a single training schedule (R+D $\rightarrow$ R or D).

During training, we randomly select modalities to drop (including no drop) at every iteration: R$\rightarrow$R+I$\rightarrow$D$\rightarrow$I$\rightarrow$$\cdots$.
We trained the whole framework including \ours in an end-to-end manner.
During testing, all testing samples of one or more modalities are dropped.

\vspace{\paramargin}\paragraph{Architecture Specifications.}
We set the default diemnsion of \ours to 512, and set the hidden dimension to 2048 in feed forward networks consisting of two linear layers and GeLU~\cite{hendrycks2016gaussian} activation.
The number of encoder and decoder layers of \ours is set to two each, and eight heads are used for self-attention operations.